\documentclass[twoside]{article}

\usepackage[accepted]{aistats2025}
\usepackage[round]{natbib}

\usepackage{booktabs}
\usepackage{graphicx}
\usepackage{caption}
\usepackage{subcaption}

\usepackage[linesnumbered, ruled, vlined]{algorithm2e}

\SetCommentSty{mycommfont}
\SetKwInput{KwInput}{Input}                
\SetKwInput{KwOutput}{Output} 
\SetKwInput{KwParams}{Parameters}  

\usepackage{amsthm}
\theoremstyle{definition}
\newtheorem{definition}{Definition}[section]
\newtheorem{theorem}{Theorem}[section]

\usepackage{amssymb}
\usepackage{mathabx}
\usepackage{hyperref}
\hypersetup{
     colorlinks   = true,
     linkcolor    = blue,
     citecolor    = blue
}

\newcommand{\vecOp}[1]{\operatorname{vec}\left(#1\right)}
\newcommand{\tenOp}[1]{\operatorname{ten}\left(#1\right)}

\newcommand{\Real}{\mathbb{R}}
\newcommand{\norm}[1]{\left\lVert#1\right\rVert} 
\newcommand{\tens}[1]{\boldsymbol{\mathcal{#1}}} 
\newcommand{\mat}[1]{\boldsymbol{#1}} 
\newcommand{\vect}[1]{\boldsymbol{#1}} 
\newcommand{\inRe}[1]{\in\Real^{#1}} 
\newcommand{\inCo}[1]{\in\mathbb{C}^{#1}} 

\begin{document}

%

%

\twocolumn[
\aistatstitle{Tensor Network Based Feature Learning Model}
\aistatsauthor{Albert Saiapin \And Kim Batselier}
\aistatsaddress{Delft Center for Systems and Control \\ Delft University of Technology \And  Delft Center for Systems and Control \\ Delft University of Technology} 
]

\begin{abstract}
Many approximations were suggested to circumvent the cubic complexity of kernel-based algorithms, allowing their application to large-scale datasets. One strategy is to consider the primal formulation of the learning problem by mapping the data to a higher-dimensional space using tensor-product structured polynomial and Fourier features. The curse of dimensionality due to these tensor-product features was effectively solved by a tensor network reparameterization of the model parameters. 
However, another important aspect of model training — identifying optimal feature hyperparameters — has not been addressed and is typically handled using the standard cross-validation approach.
In this paper, we introduce the Feature Learning (FL) model, which addresses this issue by representing tensor-product features as a learnable Canonical Polyadic Decomposition (CPD). By leveraging this CPD structure, we efficiently learn the hyperparameters associated with different features alongside the model parameters using an Alternating Least Squares (ALS) optimization method.
We prove the effectiveness of the FL model through experiments on real data of various dimensionality and scale. The results show that the FL model can be consistently trained 3-5 times faster than and have the prediction quality on par with a standard cross-validated model.
\end{abstract}

\section{INTRODUCTION}  
In the supervised learning setting, there are two main goals: to estimate a function $f_{\vect{\theta}}: \mathcal{X} \rightarrow \mathcal{Y}$ given the data $T = \{(\vect{x}_n, y_n)\}_{n=1}^{N}$, where $\vect{x}_n \in \mathcal{X}$ and $y_n \in \mathcal{Y}$, and to find the optimal model hyperparameters $\vect{\theta}$.

As a solution to the first problem, kernel methods, such as Support Vector Machines (SVMs) \citep{SVM_Cortes_1995} and Gaussian Processes (GPs)~\citep{GP_Rasmussen_2006}, were developed and have become a prominent class of machine learning techniques designed to infer non-linear functions. 
Due to a well-established mathematical framework in terms of Reproducing Kernel Hilbert Spaces~\citep{RKHS_Crowley_2021} and theoretical guarantees of convex optimization theory, kernel methods have become widely studied and applied in different problem domains. Moreover, they can be considered universal function approximators when an appropriate kernel is chosen \citep{UAT_Hammer_2003}, and recent research highlights some connections with neural networks \citep{DNN_GP_Lee_2018, BDCNN_GP_Novak_2020}. However, a major limitation of kernel machines is the need to compute the kernel matrix, which captures pairwise similarities between data points in the feature space. This computation has a complexity of at least $\mathcal{O}(N^3)$, making the method impractical for large datasets ($N \approx 10^7-10^9$) on low-performance computing systems.

One way to address this problem is to consider the primal formulation of the learning task. In this case, the data is mapped into a high-dimensional feature space, where linear inference is performed:
\begin{equation}\label{intro_simple_model}
    f_{\vect{\theta}}(\vect{x}) = \vect{\phi}_{\vect{\theta}}(\vect{x})^\top\vect{w}.
\end{equation}
The mapping $\vect{\phi}_{\vect{\theta}}(\vect{x})$ has a significant role as it allows for modeling various non-linear behaviors in the data. This paper focuses on tensor-product features, defined as:
\begin{equation}\label{tensor_product_features}
    \vect{\phi}_{\vect{\theta}}(\vect{x}) = \vect{\psi}_{\vect{\theta}}^{(D)}(x_D) \otimes \dots \otimes \vect{\psi}_{\vect{\theta}}^{(1)}(x_1),
\end{equation}
where $\vect{\psi}_{\vect{\theta}}^{(d)}: \mathbb{C} \rightarrow \mathbb{C}^{I_d}$ is a feature map acting on the $d$-th component of $\vect{x} \inCo{D}$ and $\otimes$ denotes the Kronecker product \citep{TD_Kolda_2009}. The same tensor-product structure is related to product kernels \citep{VFF_Hensman_2017, GPRR_Solin_2019}, Fourier features~\citep{TT_FF_Wahls_2014} and polynomials~\citep{KM__Cristianini_2004}. At first glance, a disadvantage of the tensor-product structure in Equation~\eqref{tensor_product_features} is that the input vector $\vect{x}$ is mapped into an exponentially large feature vector $\vect{\phi}_{\vect{\theta}}(\vect{x}) \inCo{I_1 I_2 \dots I_D}$. As a result, the model is also described by an exponential number of model parameters $\vect{w}$. This exponential scaling in the number of features limits the use of tensor-product features to low-dimensional data or to the mappings of low degree.

One way to take advantage of the existing tensor-product structure in Equation~\eqref{tensor_product_features} is by imposing a tensor network~\citep{TD_Kolda_2009, TN1_Cichocki_2016} constraint on the $\vect{w}$ parameters. For example, using a polyadic rank-$R$ constraint reduces the storage complexity of the model parameters from $\mathcal{O}(I^D)$ down to $\mathcal{O}(DIR)$ and enables the development of efficient learning algorithms with a computational complexity of $\mathcal{O}(DIR)$ per gradient descent iteration, where $I = \max(I_1, \dots, I_D)$. This idea has been explored for polynomials~\citep{FM_Rendle_2010, batselier2017tensor, PN_FM_Blondel_2017}, pure-power-1 polynomials~\citep{EM_Novikov_2017}, pure-power polynomials of higher degree~\citep{TTP_Chen_2017, QTNM_Wesel_2024} and Fourier features~\citep{TT_FF_Wahls_2014, TN_Schwab_2016, CPD_Kargas_2021, SL_Cheng_2021, TD_FF_Wesel_2021, QTNM_Wesel_2024}. In this paper, we focus on Fourier features constructed from Fourier basis functions: $\vect{\psi}_{\theta}(x)=\left[e^{-\frac{2\pi j x k}{\theta}}\right]_{k = 0}^{I_d - 1}$, where $k$ is a basis frequency and $\theta$ is a hyperparameter. In addition, $\vect{w}$ is expressed as a Canonical Polyadic Decomposition (CPD) following~\citep{TD_FF_Wesel_2021, QTNM_Wesel_2024}.

The second challenge in supervised learning - hyperparameters search - is often hardly addressed~\citep{TN_Schwab_2016, TTP_Chen_2017, SL_Cheng_2021, TD_FF_Wesel_2021, QTNM_Wesel_2024} and typically relies on application of generic cross-validation techniques~\citep{CV_Refaeilzadeh_2009}. However, these techniques do not exploit the specific structure of the model.

In this article, we introduce a new model that enables us to avoid the use of cross-validation, which is typically employed to determine the optimal value of the $\theta$ hyperparameter. 
Instead, our model considers a set of $P$ different values of $\theta$ simultaneously, using an additional tensor-product structure~\eqref{tensor_product_features}, which is leveraged to derive a computationally efficient training algorithm. Our new Feature Learning (FL) model is described by
\begin{equation*}
    f(\vect{x}) = \left[\sum_{p=1}^P \lambda_p\, \vect{\phi}_{\theta_p}(\vect{x})\right]^\top \vect{w},
\end{equation*} 
which constructs the feature map as a linear combination of $P$ different tensor-product feature maps $\vect{\phi}_{\theta_p}$. The $\lambda_p$ parameters can then be interpreted as a measure of the importance of a particular hyperparameter $\theta_p$. The linear combination of features $\vect{\phi}_{\theta_p}(\vect{x})$ is a CPD itself. A key benefit of our FL model is that the $\lambda_p$ parameters can also be efficiently learned from data. Exploiting the CPD structure of both $\vect{w}$ and the feature map allows us to improve the computational complexity of training the FL model by splitting the main non-linear optimization problem into a series of much smaller linear problems. This reduces the training computational complexity from $\mathcal{O}(I^{2D}[N + I^D])$ to $\mathcal{O}(\mathcal{E}DNIR[P + IR])$ for the large-scale high-dimensional problem, where $\mathcal{E}$ is the number of training epochs, $N$ is the number of training samples, $D$ is the dimensionality of data, $I$ is the maximum dimensionality of a feature map $\vect{\psi}_{\theta}$, $R$ is a CPD rank, $P$ is the number of different feature maps. 

Furthermore, our numerical experiments demonstrate that the FL model can be effectively applied to datasets that are large both in sample size and dimensionality. Compared to the standard cross-validation technique, the FL model trains 3–5 times faster without any significant drop in prediction performance. In addition to introducing the FL model, we also explore different forms of $\vect{\lambda}$ regularization and highlight their respective advantages, disadvantages and use cases. 

\section{BACKGROUND}
We denote scalars in italics $w, W$, vectors in non-capital bold $\vect{w}$, matrices in capital bold $\mat{W}$ and tensors, also being high-order arrays, in capital italic bold font $\tens{W}$. Sets are denoted with calligraphic capital letters - $\mathcal{Z}$. A range of natural numbers from $1$ to $N$ is denoted as $\overline{1, N}$. The $i$-th entry of a vector $\vect{w} \inCo{I}$ is denoted as $w_i$ and the $i_1i_2\dots i_D$-th entry of a $D$-order tensor $\tens{W} \inCo{I_1 \times I_2 \times \dots \times I_D}$ as $w_{i_1 i_2 \dots i_D}$. 
The conjugate-transpose of $\mat{A}$ is denoted as $\mat{A}^\top$ 
and $\otimes$, $\odot_R$, $\oasterisk$, $\oslash$ represent the Kronecker product, row-wise Khatri-Rao product, Hadamard product, element-wise division of matrices correspondingly~\citep{TD_Kolda_2009, TN1_Cichocki_2016}. Further we define two useful interconnected operators from tensor linear algebra: vectorization and tensorization.
\begin{definition}[Vectorization]
    The vectorization operator $\vecOp{\cdot}: \mathbb{C}^{I_1 \times I_2 \times \dots \times I_D} \rightarrow \mathbb{C}^{I_1I_2\dots I_D}$ such that:
    \begin{equation*}
        \vecOp{\tens{W}}_i = w_{i_1 i_2 \dots i_D},
    \end{equation*}
    where $i = i_1 + \sum_{d=2}^{D}i_d \prod_{j=1}^{d-1}I_j$.
\end{definition}

\begin{definition}[Tensorization]
    The tensorization operator $\tenOp{\cdot, I_1, I_2, \dots, I_D}: \mathbb{C}^{I_1I_2\dots I_D} \rightarrow \mathbb{C}^{I_1 \times I_2 \times \dots \times I_D}$ such that:
    \begin{equation*}
        \tenOp{\vect{w}, I_1, I_2, \dots, I_D}_{i_1 i_2 \dots i_D} = w_i.
    \end{equation*}
\end{definition}

\subsection{Tensorized Kernel Machines}
The main idea of tensor networks (TNs)~\citep{TD_Kolda_2009, TD_Cichocki_2014, TN1_Cichocki_2016} is to represent a $D$-th order tensor $\tens{W}$ as a multilinear function of small-order tensors - cores. In this work, we use the Canonical Polyadic Decomposition.
\begin{definition}[Canonical Polyadic Decomposition~\citep{TD_Kolda_2009}]
    A $D$-th order tensor $\tens{W} \in \mathbb{C}^{I_1 \times I_2 \times \dots \times I_D}$ has a rank-$R$ CPD if
\begin{equation}\label{CPD_Kron}
    \vecOp{\tens{W}} = \sum_{r=1}^{R} \vect{w}^{(D)}_{r} \otimes \dots \otimes \vect{w}^{(1)}_{r}.
\end{equation}  
\end{definition}
The cores of this tensor network are the matrices $\mat{W}^{(d)} \in \mathbb{C}^{I_d \times R}$, where $\vect{w}^{(d)}_{r}$ represents the $r$-th column of the matrix $\mat{W}^{(d)}$. In other words, the vectorized version of a tensor $\tens{W}$ can be represented as a sum of Kronecker product of vectors. Since there are $D$ matrices, the storage complexity of a $D$-th order tensor is reduced from $I^D$ down to $DIR$ by using the CPD. 
The main advantage of the CPD, in the context of this paper, is that it has only one hyperparameter, $R$, whereas the Tensor Train (TT)~\citep{TT_Oseledets_2011} decomposition has $D-1$ rank hyperparameters. Moreover, the CPD does not exhibit the exponential dependence on $D$ that is present in the Tucker Decomposition~\citep{TD_Kolda_2009}.

CPD kernel machines exploit the tensor-product structure of the features~\eqref{tensor_product_features} by restricting the model parameters $\vect{w}$ to be a tensor in the CPD format. 
This reduces the exponential complexity in the data dimensionality $D$ to a linear complexity.
\begin{theorem}[CPD Kernel Machine~\citep{QTNM_Wesel_2024}]
\label{CPD_km}
    Suppose $\tenOp{\vect{w}, I_1, I_2, \dots, I_D}$ is represented by a CPD. The model responses and gradients of
    \begin{equation*}
        f(\vect{x}) = \left[\vect{\psi}_{\theta}^{(D)}(x_D) \otimes \dots \otimes \vect{\psi}_{\theta}^{(1)}(x_1)\right]^\top \vect{w}
    \end{equation*}
    can be computed in $\mathcal{O}(DIR)$ instead of $\mathcal{O}(\prod_{d=1}^D I_d)$, where $I = \max(I_1, \dots, I_D)$.
\end{theorem}
Note that the feature map of the CPD kernel machine is a rank-1 CPD. The approach of tensorizing $\vect{w}$ and approximating the resulting tensor as a low-rank decomposition has been explored and introduced in several papers on product features and tensor networks~\citep{TT_FF_Wahls_2014, TN_Schwab_2016, batselier2017tensor, EM_Novikov_2017, SL_Cheng_2021, TD_FF_Wesel_2021, QTNM_Wesel_2024}. 
Training a CPD kernel machine means solving the following non-linear non-convex optimization problem:
\begin{equation}\label{cpd_km_opt}
    \begin{split}
        &\min_{\vect{w}}\frac{1}{2} \|\vect{y} - \mat{\Phi}\vect{w}\|_2^2 + \frac{\alpha}{2}\|\vect{w}\|_2^2 \\
        & \text{s.t. }\vect{w} = \sum_{r=1}^{R} \vect{w}^{(D)}_{r} \otimes \dots \otimes \vect{w}^{(1)}_{r},
    \end{split} 
\end{equation}
where $\vect{y} = \left[y_n\right]_{n = 1}^N$, $\vect{x}_n \inCo{D}$ - $n$-th row of data matrix $\mat{X} \inCo{N\times D}$, $\mat{\Phi} = \left[ \vect{\phi}_{\theta}(\vect{x}_n)^\top \right]_{n = 1}^N \inCo{N \times I_1 I_2\dots I_D}$, $\vect{w}^{(d)}_{r}$ is the $r$-th column of the $d$-th CPD core $\mat{W}^{(d)}$, and $\alpha$ is a regularization hyperparameter.
Note that, in order to compute the model response $\vect{f}=\mat{\Phi}\vect{w}$, we implicitly consider its real part as we generally work with complex features $\vect{\phi}_{\theta}$ and model parameters $\vect{w}$.

Several algorithms have been proposed to solve the optimization problem~\eqref{cpd_km_opt} efficiently by exploiting the CPD structure such as Alternating Least Squares (ALS)~\citep{TD_Kolda_2009}, Riemannian optimization~\citep{EM_Novikov_2017}, and first-order gradient-based optimization algorithms~\citep{ADAM_Ba_2017}.

\begin{figure*}[!ht]
\vspace{.3in}
    \centering
     \begin{subfigure}[!t]{0.4\textwidth}
         \centering
         \includegraphics[width=\textwidth]{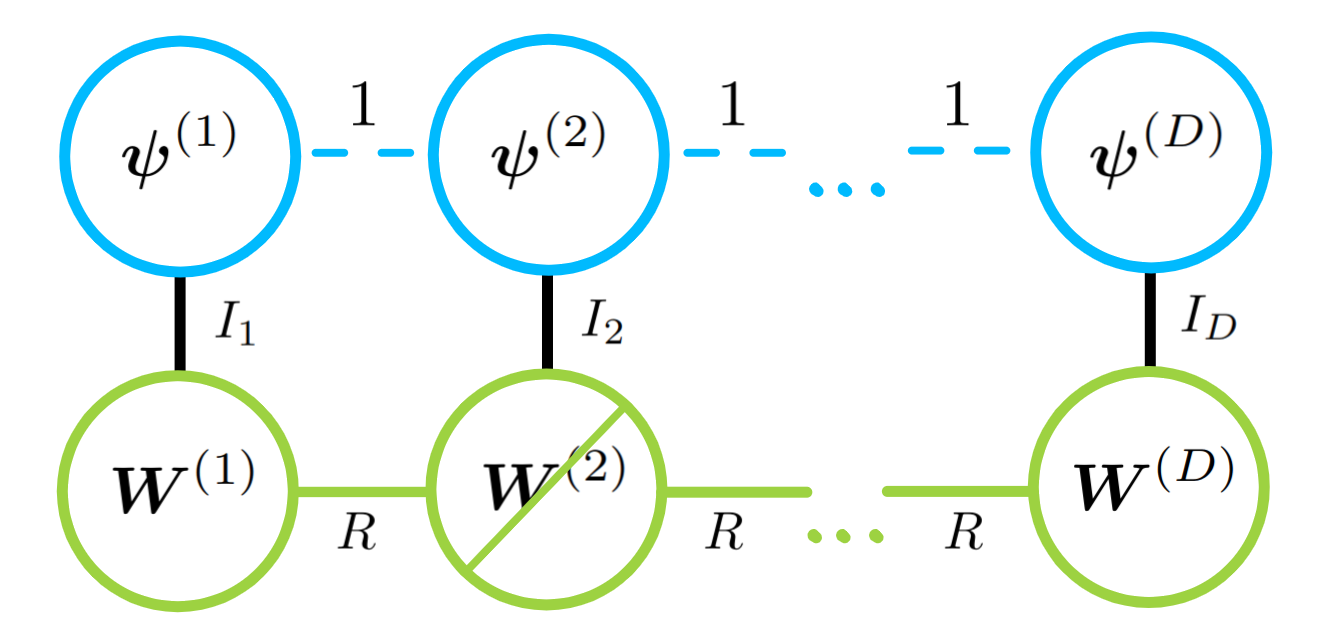}
         \caption{Tensorized CPD Kernel Machine}
         \label{fig:basic_model}
     \end{subfigure}
     \hfill
     \begin{subfigure}[!t]{0.5\textwidth}
         \centering
         \includegraphics[width=\textwidth]{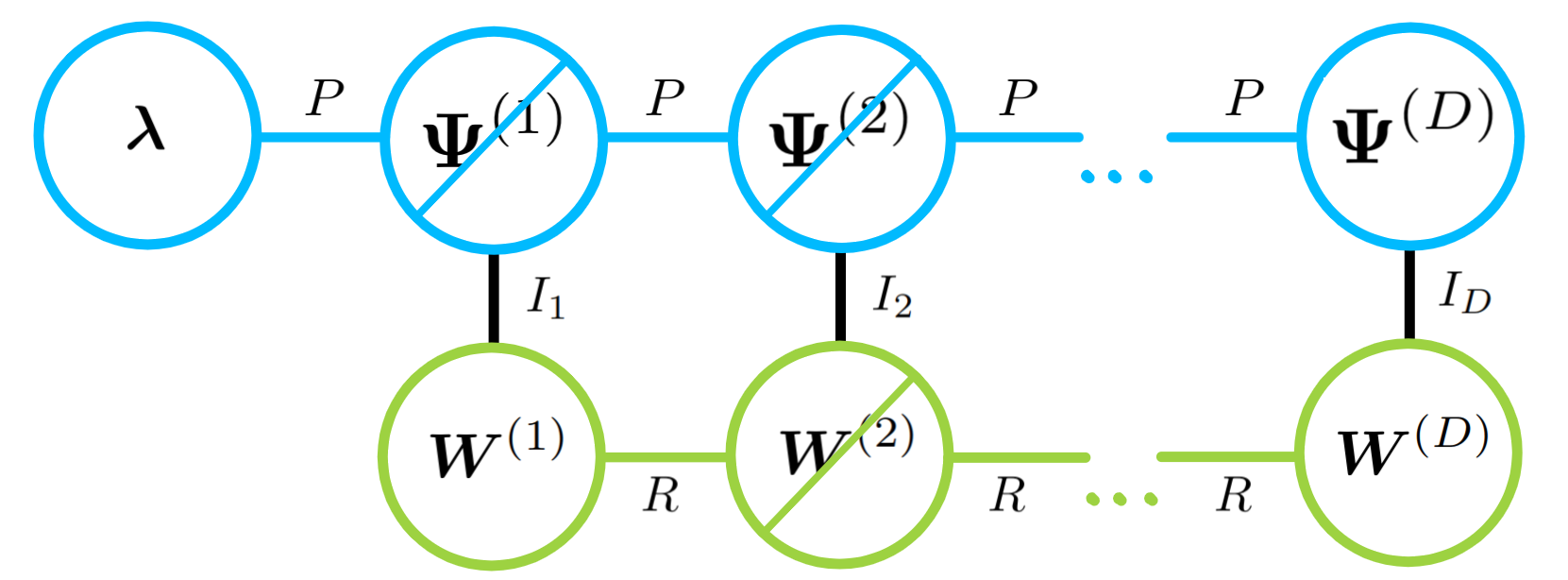}
         \caption{Feature Learning Model}
         \label{fig:FL_model}
     \end{subfigure}
\vspace{.3in}
    \caption{Tensorized CPD Kernel Machine (Figure~\ref{fig:basic_model}) and FL Model (Figure~\ref{fig:FL_model}). In the diagrams, each circle without a cross represents a vector or a matrix (defined by the number of outgoing solid lines, one or two respectively); a crossed circle depicts a three-dimensional tensor containing a particular matrix in the diagonal slice; blue color represents parameters related to non-linear features $\vect{\psi}_{\theta}^{(d)}(x_d)$, $d = \overline{1, D}$; green color represents model parameters $\vect{w}$ in a CPD format; a solid line denotes a summation along the corresponding index, while a dashed line denotes a Kronecker product~\citep{TN1_Cichocki_2016}; $\mat{\Psi}^{(d)} = [\vect{\psi}_{\theta_1}^{(d)}(x_d), \dots, \vect{\psi}_{\theta_P}^{(d)}(x_d)] \inCo{I_d \times P}$. Figure~\ref{fig:basic_model} depicts model~\eqref{intro_simple_model} with a rank-1 CPD feature map, while Figure~\ref{fig:FL_model} represents our FL model~\eqref{fl_model} with a rank-$P$ feature map. }
    \label{fig:FL_Diagram}
\end{figure*}

\subsection{Fourier Features}
Different possibilities have been explored for constructing the feature mapping $\vect{\phi}_{\theta}$, for example polynomial functions~\citep{PN_FM_Blondel_2017}, pure-power polynomials of higher degree~\citep{TTP_Chen_2017} and Fourier features~\citep{TT_FF_Wahls_2014, SL_Cheng_2021}.
Since polynomial features can exhibit instabilities during training due to the construction of ill-conditioned Vandermonde matrices, we will focus on Fourier features moving forward. However, it is important to note that other generic functions can be used for high-dimensional data mapping in the context of this work.
\begin{definition}[Fourier Features~\citep{QTNM_Wesel_2024}]
    For an input sample $\vect{x} \in \mathbb{C}^D$, the Fourier feature map $\vect{\phi}_{\theta}: \mathbb{C}^D \rightarrow \mathbb{C}^{I_1 I_2 \dots I_D}$ with $I_d$ basis frequencies $\dfrac{-I_d}{2}, \dots, \dfrac{I_d}{2} - 1$ per dimension is defined as:
    \begin{equation*}\label{ff_map}
        \begin{split}
            &\vect{\phi}_{\theta}(\vect{x}) = 
            \bigotimes_{d=1}^D c_d \;\vect{\psi}_{\theta}^{(d)}(x_d), \\
            &\vect{\psi}_{\theta}^{(d)}(x_d) = \left[e^{-\frac{2\pi j x_d k}{\theta}}\right]_{k = 0}^{I_d - 1},
        \end{split} 
    \end{equation*}
    where $j$ is the imaginary unit, $c_d = e^{2\pi j x_d \frac{2+I_d}{2\theta}}$ and $\theta$ is the periodicity hyperparameter of the function class.
\end{definition}
Applications of this mapping can be found in the field of kernel machines, as they can be viewed as eigenfunctions of a $D$-dimensional stationary product kernel~\citep{GP_Rasmussen_2006, GPRR_Solin_2019}. Generally, the corresponding optimal value of the hyperparamter $\theta$ can be identified through cross-validation across different options $\theta_1, \dots, \theta_P$. To achieve this, one would need to solve the optimization problem in~\eqref{cpd_km_opt} at least $P$ times.

In the next section, we introduce the main contribution of this paper. Our new Feature Learning model is based on the CPD kernel machine concept but, in contrast, learns a linear combination of feature maps from the data.
This model allows the use of any non-linear features with a Kronecker product structure and solves an optimization problem analogous to~\eqref{cpd_km_opt} only once.

\section{TENSOR NETWORK BASED FEATURE LEARNING MODEL}
Previous works~\citep{EM_Novikov_2017, TD_FF_Wesel_2021, QTNM_Wesel_2024} introduced similar formulations of the optimization problem~\eqref{cpd_km_opt}, with an emphasis on finding optimal model parameters - $\vect{w}$.
However, the search for model hyperparameters, $\theta$, has received little attention. A common approach is to apply cross-validation over a set of hyperparameter options, $\theta_1, \theta_2, \dots, \theta_P$. While effective for a broad range of problems, this generic technique does not leverage problem-specific characteristics that could enhance model properties, such as reducing computational cost and memory consumption or improving prediction quality. In the following subsections, we propose an alternative approach for optimizing model parameters related to the feature map $\vect{\phi}_{\theta}$ in the context of tensorized kernel machines.

\subsection{Feature Learning Model}
Instead of using the model~\eqref{intro_simple_model} with a feature map $\vect{\phi}_{\theta}$ depending on hyperparameter $\theta$, we consider the following model formulation:
\begin{equation}\label{fl_model}
    f(\vect{x}) = \left[\sum_{p=1}^P \lambda_p \,\vect{\phi}_{\theta_p}(\vect{x})\right]^\top \vect{w},
\end{equation}
with $\vect{\phi}_{\theta_p}(\vect{x}) = \vect{\psi}_{\theta_p}^{(D)}(x_D) \otimes \dots \otimes \vect{\psi}_{\theta_p}^{(1)}(x_1)$ and learnable parameters $\lambda_p$ that represent the scaling of features in the sum. The new feature mapping in the brackets can now be interpreted as a learnable rank-$P$ CPD as it is the sum of $P$ tensor-product features~\eqref{CPD_Kron}. The model reformulation allows us to replace the hyperparameters search problem with a regular optimization problem. 
In this case, we have to find two distinct sets of parameters: model parameters $\vect{w}$ and feature parameters $\vect{\lambda} = [\lambda_1, \dots,\lambda_P]^\top$. Figure~\ref{fig:FL_Diagram} demonstrates the difference between the original model~\eqref{intro_simple_model} and the FL model~\eqref{fl_model} in a tensor network diagram. For both models the lower green network represents a rank-$R$ CPD of $\tenOp{\vect{w}, I_1, I_2, \dots, I_D}$. The main difference is in the upper blue network. In the Figure~\ref{fig:basic_model}, the blue network represents the Kronecker product of vectors $\vect{\psi}_{\theta}^{(1)}, \dots, \vect{\psi}_{\theta}^{(D)}$, whereas the blue network in the Figure~\ref{fig:FL_model} depicts the rank-$P$ CPD of the feature map. 

Under the supervised regression problem framework, both $\vect{w}$ and $\vect{\lambda}$ are found by optimizing the following non-convex non-linear optimization problem:
\begin{equation}\label{fl_objective}
    \begin{split}
        &\min_{\vect{\lambda}, \vect{w}}\frac{1}{2} \|\vect{y} - \sum_{p = 1}^P\lambda_p \mat{\Phi}_p\vect{w}\|_2^2 + \frac{\alpha}{2}\|\vect{w}\|_2^2 + \beta\operatorname{Reg}(\vect{\lambda}) \\
        &\text{ s.t. }\vect{w} = \sum_{r=1}^{R} \vect{w}^{(D)}_{r} \otimes \dots \otimes \vect{w}^{(1)}_{r},
    \end{split} 
\end{equation}
where $\mat{\Phi}_p = \left[\vect{\phi}_{\theta_p}(\vect{x}_n)^\top \right]_{n = 1}^N \inCo{N \times I_1 I_2\dots I_D}$, $\vect{w}^{(d)}_{r}$ is the $r$-th column of the $d$-th CPD core $\mat{W}^{(d)}$. In contrast to the optimization problem~\eqref{cpd_km_opt} there is now an additional regularization term $\operatorname{Reg}(\vect{\lambda})$ for $\vect{\lambda}$ together with its own $\beta$ regularization hyperparameter.

In the next subsections we derive the ALS solution to the problem~\eqref{fl_objective} separating the distinct parameters $\vect{\lambda}, \mat{W}^{(1)}, \dots, \mat{W}^{(D)}$ such that each optimization sub-problem becomes linear. The local convergence of the ALS optimization using CPD has been investigated in~\citep{ALS_Uschmajew_2012}.

\subsection{ALS Update Of Model Parameters $\mat{W}^{(d)}$}
To find the optimal model parameters $\vect{w}$ represented as a CPD tensor we fix the values of $\vect{\lambda}$. Secondly, as the CPD is a multi-linear function of its cores~\citep{TN1_Cichocki_2016} we fix the values of all the cores, but one - $\mat{W}^{(d)}$. This results in the following Regularized Least Squares problem:
\begin{equation}\label{wk_update}
    \begin{split}
     &\min_{\vect{v}} \frac{1}{2} \norm{\vect{y} - \mat{A}_d \vect{v}}_2^2 + \frac{\alpha}{2}\vect{v}^\top(\mat{H}_d  \otimes \mat{I})\vect{v} + \text{const} \\
     &\text{ s.t. }\vect{v} = \operatorname{vec}(\mat{W}^{(d)}),
    \end{split}  
\end{equation}
where: 
\begin{equation}\label{wk_update_specs}
    \begin{split}
        &\mat{A}_d = \sum_{p = 1}^P\lambda_p \left(\mat{Z}^{(d, p)} \odot_R \mat{\Psi}^{(d, p)}\right) \inCo{N \times I_dR}, \\
        &\mat{H}_d = \bigoasterisk_{\substack{j=1 \\ j\neq d}}^D\mat{W}^{(j) \top} \mat{W}^{(j)} \inCo{R \times R},\\
        &\mat{Z}^{(d, p)} = \bigoasterisk_{\substack{j=1 \\ j\neq d}}^D \mat{\Psi}^{(j, p)} \mat{W}^{(j)} \inCo{N \times R}, \\
        &\mat{\Psi}^{(d, p)} = \left[\vect{\psi}_{\theta_p}^{(d)}(x_{nd})^\top \right]_{n=1}^N \inCo{N \times I_d}.
    \end{split}  
\end{equation}
In order to update the full vector of model parameters $\vect{w}$, one sequentially updates the $\mat{W}^{(d)}$, $d=\overline{1, D}$ cores of the CPD using the following expression:
\begin{equation}\label{wk_sol}
    \operatorname{vec}(\mat{W}^{(d)}) = (\mat{A}_d^\top \mat{A}_d + \alpha (\mat{H}_d  \otimes \mat{I}))^{-1}\mat{A}_d^\top\vect{y}.
\end{equation}
The computational complexity of this update for a single core is $\mathcal{O}\left(N[I_dRP + I_d^2R^2]\right)$ when $N \gg I_dR$, and the memory complexity is $\mathcal{O}(NI_dR)$. Here, $N$ is the training sample size, $I_d$ is the dimensionality of the mapping $\vect{\psi}_{\theta_p}^{(d)}$, $R$ is the CPD rank, $P$ is the number of unique feature hyperparameter configurations $\theta$.
Once all $D$ core matrices in the CPD representation of $\vect{w}$ have been updated, the model parameters $\vect{w}$ are fixed, and the next step is to update the feature parameters $\vect{\lambda}$.

\subsection{ALS Update Of Feature Parameters $\vect{\lambda}$}
The FL model objective~\eqref{fl_objective} can be expressed as a function of $\vect{\lambda}$ if the vector $\vect{w}$ vector is assumed to be fixed, resulting in a Regularized Least Squares problem:
\begin{align}\label{lambda_opt} 
    \min_{\vect{\lambda}} \frac{1}{2} \|\vect{y} - \mat{F}\vect{\lambda}\|_2^2 + \beta\operatorname{Reg}(\vect{\lambda}) + \text{const}, 
\end{align}  
where $\mat{F} = \left[ \mat{\Phi}_1\vect{w} \dots \mat{\Phi}_P\vect{w} \right] \inRe{N \times P}$.
The optimal solution depends on the type of regularization and parameter constraints. In this work, we consider 3 types of regularization: L1~\citep{Fista_Beck_2009}, L2~\citep{Ridge_Kennard_2000}, and Fixed Norm~\citep{MC_Golub_2013}.

\paragraph{L1 Regularization.}
The solution of this convex, yet not differentiable problem can be found in iterative way using proximal gradient based algorithm~\citep{Fista_Beck_2009}:
\begin{equation}\label{lambda_l1}
    \begin{split}
     &\vect{p}_{s} = \vect{\lambda}_{s} - t_s\mat{F}^\top(\mat{F}\vect{\lambda}_{s} - \vect{y}), \\
     &\vect{\lambda}_{s+1} = \operatorname{\textsc{sign}}(\vect{p}_{s}) \oasterisk \operatorname{\textsc{max}}(0, \vect{p}_{s} - \beta t_s \vect{1}), \\
     &t_s \in \left( 0, \frac{1}{\norm{\mat{F}^\top \mat{F}}_2} \right), s = \overline{1, S}, \\
    \end{split}
\end{equation}
where $\vect{\lambda}_{s}$ represents the estimate after $s$ gradient step iterations, $t_s$ is the gradient step size, $\operatorname{\textsc{sign}}$ and $\operatorname{\textsc{max}}$ are operators of element-wise sign and max functions respectively, $\vect{1}$ is a vector of all ones. The computational complexity of this algorithm is $\mathcal{O}(NP^2)$, assuming $N \gg S$.
 
Due to the nature of the L1 regularization constraint, some coefficients $\lambda_p$ become exactly zero, leading to sparse models. This sparsity can be utilized to reduce the computational complexity of the $\mat{W}^{(d)}$ update by avoiding the inclusion of zero contributions when constructing the $\mat{A}_d$ matrix in Equation~\eqref{wk_update_specs}.

\paragraph{L2 Regularization.}
Similar to the $\mat{W}^{(d)}$ update, the solution of the Least Squares problem for $\vect{\lambda}$ (Equation~\eqref{lambda_opt}) with L2 regularization is:
\begin{equation}\label{lambda_l2}
    \vect{\lambda} = (\mat{F}^\top \mat{F} + \beta \mat{I})^{-1}\mat{F}^\top\vect{y}
\end{equation}
with a computational complexity of $\mathcal{O}(NP^2)$.

\paragraph{Fixed Norm Regularization.}
The introduction of an additional $\beta$ hyperparameter can be seen as a drawback, as it requires extra tuning. One way to eliminate this hyperparameter from the optimization problem is by considering the following constrained optimization problem:
\begin{equation}\label{fixed_lambda_opt}
        \min_{\norm{\vect{\lambda}}_2^2 \leq 1} \frac{1}{2} \|\vect{y} - \mat{F}\vect{\lambda}\|_2^2. 
\end{equation}
By constraining the norm of $\vect{\lambda}$, one can effectively regularize the optimization problem. According to~\citep{MC_Golub_2013} the optimal solution of problem~\eqref{fixed_lambda_opt} is:
\begin{equation}\label{fixed_lambda_sol}
    \vect{\lambda} = \mat{V}\left(\mat{\Sigma}^\top\mat{\Sigma} + \mu\mat{I}\right)^{-1}\mat{\Sigma}^\top\vect{c},
\end{equation}
where $\mat{F} = \mat{U}\mat{\Sigma}\mat{V}^\top$ is the Singular Value Decomposition (SVD) of the matrix $\mat{F}$, $\vect{c} = \mat{U}^\top\vect{y}$, and $\mu$ is the solution of the non-linear equation: 
\begin{equation*}
    \sum_{i=1}^r \frac{c_i^2 \sigma_i^2}{(\sigma_i^2 + \mu)} = 1.
\end{equation*}
As the SVD is the dominant computational part of this algorithm, its complexity is $\mathcal{O}(NP^2)$.

{\setlength{\algomargin}{1.5em}
\begin{algorithm}[!t]
\DontPrintSemicolon
  \KwParams{$\alpha, \beta, R, \mathcal{E}, [\vect{\psi}_{\theta_p}^{(d)}]_{d=1, p=1}^{D, P}$}
  \KwData{$\mat{X} \inCo{N\times D}, \vect{y} \inRe{N}$}
  Initialize $\mat{W}^{(1)}, \dots, \mat{W}^{(D)}, \vect{\lambda}$\\
  Precompute $\mat{Z}_p = \bigoasterisk_{\substack{j=1}}^D \mat{\Psi}^{(j, p)} \mat{W}^{(j)}, p=\overline{1, P}$ \\
  Precompute $\mat{H} = \bigoasterisk_{\substack{j=1}}^D \mat{W}^{(j) \top} \mat{W}^{(j)}$ \\
  \While{$e \leq \mathcal{E}$}
   {
        \For{$k = \overline{1, D}$}{
            $\mat{Z}^{(k, p)} = \mat{Z}_p \oslash [\mat{\Psi}^{(k, p)} \mat{W}^{(k)}] , p=\overline{1, P}$ \\
            $\mat{H}_k = \mat{H} \oslash [\mat{W}^{(k) \top} \mat{W}^{(k)}]$\\
            Update $\mat{A}_k$ from~\eqref{wk_update_specs} based on $\mat{Z}^{(k, p)}$ \\
            Update $\mat{W}^{(k)}$ based on~\eqref{wk_sol}\;
            $\mat{Z}_p = \mat{Z}^{(k, p)}\oasterisk [\mat{\Psi}^{(k, p)} \mat{W}^{(k)}] , p=\overline{1, P}$ \\
            $\mat{H} = \mat{H}_k \oasterisk [\mat{W}^{(k) \top} \mat{W}^{(k)}]$\\
        }
   		Update $\vect{\lambda}$ based on~\eqref{lambda_l1} or~\eqref{lambda_l2} or~\eqref{fixed_lambda_sol}\;
        $e = e + 1$\;
   }
   \KwOutput{$\mat{W}^{(1)}, \dots, \mat{W}^{(D)}, \vect{\lambda}$}
\caption{FL Model ALS Algorithm}
\label{fl_alg}
\end{algorithm}
}
\subsection{FL Model Implementation Details}

As we mentioned earlier, the optimization problem~\eqref{fl_objective} is effectively solved using an Alternating Least Squares approach for $\vect{\lambda}$ and $\vect{w}$. The parameters are updated over multiple epochs or until convergence, based on a predefined metric or loss function. Implementation details are provided in Algorithm~\ref{fl_alg}.

In line 1 of Algorithm~\ref{fl_alg}, the FL model parameters are randomly initialized. The initial CPD cores $\mat{W}^{(d)}, d=\overline{1,D}$ are sampled from a standard normal distribution and then normalized column-wise. The initial feature parameters, $\vect{\lambda}$, are drawn from a uniform distribution from 0 to 1. After the initialization step, computational costs can be reduced by precomputing the matrices from lines 2-3 in Algorithm~\ref{fl_alg}. This precomputation has to be done once, before the training loop starts.
The remaining lines in Algorithm~\ref{fl_alg} are the repeated updates of every core $\mat{W}^{(d)}, d=\overline{1, D}$ and feature parameters $\vect{\lambda}$. The overall computational complexity of the FL model training is $\mathcal{O}(\mathcal{E}DNIR[P + IR])$ and peak memory complexity is $\mathcal{O}(NR[P+I])$, computed based on Equations~\eqref{wk_sol},~\eqref{lambda_l1},~\eqref{lambda_l2},~\eqref{fixed_lambda_sol} and Algorithm~\ref{fl_alg}.

It is worth mentioning that Fourier features (Definition~\ref{ff_map}) can be \textit{quantized}, meaning further tensorized~\citep{QTNM_Wesel_2024}. These quantized features allow for the quantization of the model weights $\vect{w}$ yielding more expressive models for the same number of model parameters. For example, suppose the dimensionality $I_d$ of $\vect{\psi}_{\theta}^{(d)}$ in Equation~\eqref{tensor_product_features} is $I_d = 2^{K_d}$. Then each $\vect{\psi}_{\theta}^{(d)}$ itself can be written as a Kronecker product
\begin{equation*}
    \vect{\psi}_{\theta}^{(d)} = \vect{\gamma}^{(K_d)}_{\theta} \otimes \dots \otimes \vect{\gamma}^{(1)}_{\theta},
\end{equation*}
where $\vect{\gamma}^{(k)}_{\theta} \in \mathbb{C}^{2}, k=\overline{1, K_d}$. In this case, CPD kernel machine consists of $\sum_{d=1}^D K_d$ cores instead of $D$, with each core $\mat{W}^{(d)}$ of size $2 \times R$ instead of $I_d \times R$.

As the difference between the two tensorization methods is negligible in terms of notation and design, our implementation uses the quantized version as it shows better generalization capabilities~\citep{QTNM_Wesel_2024}. In the quantized setting the FL model computational complexity is $\mathcal{O}(\mathcal{E}DNKR[P + R])$ and the corresponding peak memory complexity is $\mathcal{O}(NRP)$ with $K = \log_2(I)$, $I = \max(I_1, \dots, I_D)$. As a result, the quantized algorithm requires less memory for intermediate calculations and less time to update the model parameters.

\begin{table*}[!ht]
\caption{Impact of regularization on the FL model performance across various datasets. MSE and Time metric values were averaged over 10 restarts. Reg. column denotes different regularization types: L1 represents $\norm{\vect{\lambda}}_1$, L2 denotes $\norm{\vect{\lambda}}_2$, FN enforces the constraint $\norm{\vect{\lambda}}_2 \leq 1$. Option (P) introduces an additional constraint to the problem, requiring the solution to be non-negative.}
\label{table_comparison_configs}
\begin{center}
\begin{tabular}{||l||c|c|c|c|c||c|c|c|c|c||}
\toprule
 & \multicolumn{5}{c||}{MSE $\downarrow$} & \multicolumn{5}{c||}{Time (sec.) $\downarrow$} \\
 \midrule
Reg. & Airfoil & Concrete & Energy & Wine & Yacht & Airfoil & Concrete & Energy & Wine & Yacht \\
\midrule
FN & 0.19 & 0.154 & 0.003 & 0.68 & 0.11 & 4.9 & 1.3 & 0.7 & 26.9 & 0.1 \\
FN(P) & 0.187 & 0.14 & 0.009 & 0.932 & 0.366 & 4.1 & 1.3 & 0.7 & 23.9 & 0.1 \\
L1 & 0.184 & 0.139 & 0.003 & 0.692 & 0.112 & 2.7 & 1.4 & 1.0 & 24.9 & 0.1 \\
L1(P) & 0.182 & 0.176 & 0.003 & 0.705 & 0.115 & 3.4 & 1.3 & 1.5 & 23.7 & 0.1 \\
L2 & 0.189 & 0.15 & 0.003 & 0.776 & 0.1 & 3.1 & 1.5 & 0.7 & 27.8 & 0.1 \\
L2(P) & 0.188 & 0.146 & 0.007 & 0.672 & 0.327 & 3.4 & 1.4 & 0.7 & 26.8 & 0.1 \\
\bottomrule
\end{tabular}
\end{center}
\end{table*}

\section{NUMERICAL EXPERIMENTS}\label{experiments}
In all experiments the input $\vect{x}$ is scaled to lie in the
unit hypercube and the output vector $\vect{y}$ is standardized. We consider quantized version of Fourier features (Definition~\ref{ff_map}) for all the experiments.
The model parameters $\vect{w}$ are represented as a quantized CPD of rank $R$. For the cross-validation baseline (CV), we use a CPD kernel machine (Theorem~\ref{CPD_km}, \cite{QTNM_Wesel_2024}) with quantization and apply 6-fold cross-validation to the $\theta$ hyperparameter of the Fourier features, with $\theta \in \{10, 2, 128, 25, 64, 600, 2000, 1024\}$. For the experiments, we select 5 UCI regression datasets~\citep{UCI}. For each dataset, 80\% of the data is randomly chosen for training, with the remaining 20\% reserved for testing. We set the same number of basis functions, $I_d = I$, uniformly for all $d=\overline{1, D}$ for simplicity. The CPD rank $R$ and hyperparameter $I$ are chosen such that the number of FL model parameters is less than the sample size $N$, ensuring an underparameterized training regime to prevent the data memorization.
Algorithm~\ref{fl_alg} is performed for 10 full updates (epochs) of $\vect{w}$ and $\vect{\lambda}$. The training procedure is repeated 10 times, with the dataset split into training and test sets as described above. Finally, we compute the mean and standard deviation of the mean squared error (MSE) on a test set and check training time (in seconds). All experiments, except for the large-scale one, were conducted on a Dell Inc. Latitude 7440 laptop equipped with a 13th Gen Intel Core i7-1365U CPU and 16 GB of RAM.

The source code for all the functions and classes, written in Python, and data to reproduce all experiments is available on GitHub.\footnote{\url{https://github.com/AlbMLpy/TN-FL-Model}}

\begin{table*}[!ht]
    \centering
    \caption{Test MSE results and training time of the FL model and corresponding Cross-Validated model. For all the datasets $P=8$ and the results were calculated using 10 restarts. For Airline data, $P=6$ and the results were calculated based on 5 restarts. MSE (FL)/(CV) - the average test MSE value within 1 std; Time (FL)/(CV) - the average training time (in seconds);  $N$ - sample size; $D$ - data dimensionality; $I$ - the number of basis functions per dimension; $R$ - the CPD rank.}
    \label{table:cv-fl-comparison}
    \begin{center}
    \scalebox{1.0}{\begin{tabular}{||l||c|c|c|c||c|c|c|c||}
    \toprule
    Data & $N$ & $D$ & $I$ & $R$ & MSE (FL) $\downarrow$& MSE (CV) $\downarrow$& Time (FL) $\downarrow$& Time (CV) $\downarrow$\\
    \midrule
    Airfoil & 1502 & 5 & 4 & 51 & 0.184 $\pm$ 0.02 & 0.223 $\pm$ 0.02 & 3.0 & 23 \\
    Energy & 768 & 8 & 4 & 15 & 0.003 $\pm$ 0.0 & 0.003 $\pm$ 0.0 & 0.91 & 5.5 \\
    Yacht & 308 & 6 & 2 & 6 & 0.112 $\pm$ 0.02 & 0.358 $\pm$ 0.06 & 0.149 & 0.615 \\
    Concrete & 1030 & 8 & 8 & 10 & 0.139 $\pm$ 0.03 & 0.118 $\pm$ 0.02 & 1.2 & 8.8 \\
    Wine & 6497 & 11 & 16 & 25 & 0.692 $\pm$ 0.07 & 0.652 $\pm$ 0.04 & 33 & 152 \\
    \midrule
    Airline & 5929413 & 8 & 64 & 20 & 0.804 $\pm$ 0.0 & 0.779 $\pm$ 0.0 & 15159 & 56590 \\
    \bottomrule
    \end{tabular}}
    \end{center}
\end{table*}

\subsection{Regularization Alternatives}

The primary goal of the first set of experiments was to examine how different regularization terms, $\operatorname{Reg}(\vect{\lambda})$, in Equation~\eqref{fl_objective} affect the predictive quality of the regression solution. We consider 3 types of regularization: L1 - $\norm{\vect{\lambda}}_1$, L2 - $\norm{\vect{\lambda}}_2$, Fixed Norm is $\norm{\vect{\lambda}}_2 \leq 1$. Apart from that, we also consider an additional non-negative constraint on the final solution - $\lambda_p \geq 0$. As a result, there are 6 cases to examine.

Table~\ref{table_comparison_configs} demonstrates that different regularization functions yield comparable results in terms of prediction quality (test MSE) and training time across all datasets. However, the FL model combined with L1 and L1(P) settings achieved slightly better MSE values and trained faster on average across different datasets. An intuitive explanation for this result is that the computational complexity of the L1 solution is reduced due to sparsity. Specifically, a fraction of the feature parameters $\lambda_p = 0$, which results in fewer computations to update matrices $\mat{W}^{(d)}\text{, } d=\overline{1, D}$ in Equation~\eqref{wk_update}.

To summarize the results of Table~\ref{table_comparison_configs}, differently regularized problems yield close solutions in terms of the MSE metric and training time. Therefore one should look for other criteria to distinguish the different regularization methods. For example, L1 regularization generates a sparse solution and the more $\lambda_p = 0$, the faster the ALS method would work for the whole FL model due to reduced computations. However, one might need to do cross-validation with respect to $\beta$ - regularization hyperparameter. On the other hand, fixed norm regularization allows to avoid extra hyperparameter tuning, but the solution would be dense in terms of $\vect{\lambda}$ in this case. 

\subsection{Comparison With Cross-Validation}
In this section we compare the FL model training abilities to the corresponding quantized CPD kernel machine~\citep{TD_FF_Wesel_2021, QTNM_Wesel_2024} with cross-validation (CV) in the sequential learning setting (parallelization is turned off for fair comparison). 
In order to do that, we train these algorithms on several small- and large-scale datasets and showcase the test MSE and training time.

\begin{figure*}[h]
\vspace{.3in}
\includegraphics[width=170mm]{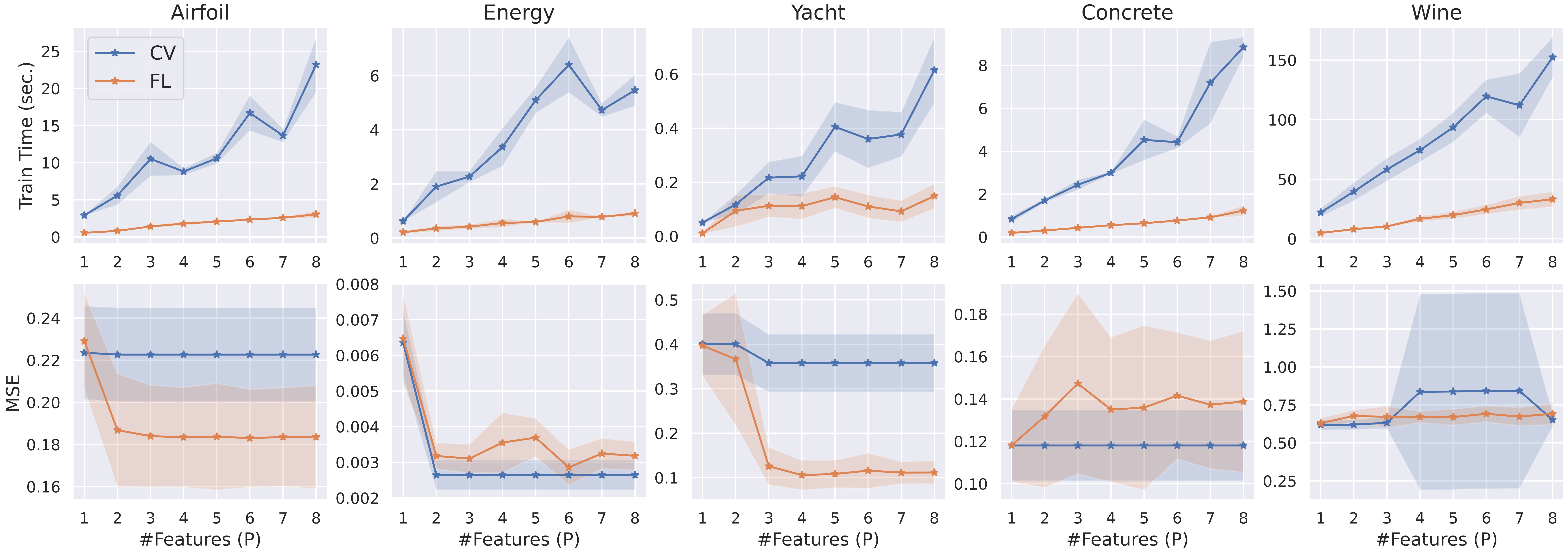}
\vspace{.3in}
\caption{
Plots of the training time (first row) and test MSE (second row)  of FL and CV models (orange and blue curves respectively) as a function of the number of features $P$ for different real-life datasets (column-wise). Solid lines represent mean metric calculations and shaded regions depict $\pm 1$ standard deviation around the mean across 10 restarts. The proposed FL model requires consistently less time to train compared to the conventional cross-validation. Likewise the prediction error of the FL model is either similar to CV (shaded regions intersect) or significantly lower (Yacht data) that demonstrates the superiority of the FL model.
}
\label{dynamics_fl}
\end{figure*}

\subsubsection{Small-Scale Problems}
Figure~\ref{dynamics_fl} shows how training time and test MSE are affected by the number of features $P$ (i.e., different number of hyperparameter options). It can be observed that the FL model exhibits a less steep learning curve in terms of training time for all datasets. This is due to the training algorithm structure and the choice of L1 regularization that encourages sparser representation of features, thereby reducing the computational load between epochs. The total computational complexity of the FL model: $\mathcal{O}(\mathcal{E}DNIR[P + IR])$ is better than that of CV: $\mathcal{O}(\mathcal{E}DNIR[PIR])$.
Speaking of the prediction error, we can conclude that the FL model shows performance similar to CV in most datasets (mean values are close and standard deviation regions intersect). If we consider Yacht dataset the quality of the FL model is superior to CV in that case. A possible explanation for this result is that the sum of features (FL) appears to be more expressive than one particular feature (CV). 

In summary, according to Table~\ref{table:cv-fl-comparison} and Figure~\ref{dynamics_fl}, the FL model can be trained 3-5 times faster without a significant drop in prediction quality compared to cross-validation.

\subsubsection{Large-Scale Problem}
In order to show and explore the behavior of the FL model on large scale data, we consider the airline dataset~\citep{hensman2013gaussianprocessesbigdata}, an 8-dimensional dataset which consists of $N = 5929413$ recordings of commercial airplane flight delays that occurred in 2008 in the USA. Specifically for this dataset~\citep{samo2016stringmembranegaussianprocesses} we consider a uniform random draw of $2/3N$ for training and keep the remainder for the evaluation of MSE on the test set and repeat the procedure 5 times. We use Fourier features with local dimension $I_d = 64$. For this experiment, we set $\alpha=0.01$, $\beta=0.1$, $R=20$, and run the ALS optimizer for 10 epochs. The different options for $\theta$ are $\{10, 2, 128, 25, 64, 1024\}$. This experiment was conducted on the 2 * AMD EPYC 7252 8-core CPU, 256 GB RAM memory.

As seen from the last row of Table~\ref{table:cv-fl-comparison}, training the FL model took four times less time, with a non-significant difference in prediction quality (0.80 compared to 0.78). The results of this experiment further confirm that the unique structure of the FL model enhances its efficiency.

\section{CONCLUSION}
We introduced a new Feature Learning (FL) model that uses the CPD structure for features representation and model parameters $\vect{w}$. We verified experimentally the effectiveness of the FL model on real small and large scale datasets, and demonstrated that it can be trained consistently faster than the cross-validated CPD kernel machine.
For future work, we plan to develop a parallel FL model utilizing the summation structure of the model. Another promising direction is to explore a probabilistic formulation of the FL model, enabling uncertainty estimation for classification and regression tasks.

\subsubsection*{Acknowledgements}
We would like to thank the anonymous reviewers for their numerous suggestions and improvements which have greatly improved the quality of this paper.
This publication is part of the project Sustainable learning for Artificial Intelligence from noisy large-scale data (with project number VI.Vidi.213.017) which is financed by the Dutch Research Council (NWO).

\bibliography{Bibfile}

\section*{Checklist}


 \begin{enumerate}

 \item For all models and algorithms presented, check if you include:
 \begin{enumerate}
   \item A clear description of the mathematical setting, assumptions, algorithm, and/or model. Yes
   \item An analysis of the properties and complexity (time, space, sample size) of any algorithm. Yes
   \item (Optional) Anonymized source code, with specification of all dependencies, including external libraries. Yes
 \end{enumerate}

 \item For any theoretical claim, check if you include:
 \begin{enumerate}
   \item Statements of the full set of assumptions of all theoretical results. Yes
   \item Complete proofs of all theoretical results. Yes
   \item Clear explanations of any assumptions. Yes     
 \end{enumerate}

 \item For all figures and tables that present empirical results, check if you include:
 \begin{enumerate}
   \item The code, data, and instructions needed to reproduce the main experimental results (either in the supplemental material or as a URL). Yes
   \item All the training details (e.g., data splits, hyperparameters, how they were chosen). Yes
         \item A clear definition of the specific measure or statistics and error bars (e.g., with respect to the random seed after running experiments multiple times). Yes
         \item A description of the computing infrastructure used. (e.g., type of GPUs, internal cluster, or cloud provider). Yes
 \end{enumerate}

 \item If you are using existing assets (e.g., code, data, models) or curating/releasing new assets, check if you include:
 \begin{enumerate}
   \item Citations of the creator If your work uses existing assets. Yes
   \item The license information of the assets, if applicable. Not Applicable
   \item New assets either in the supplemental material or as a URL, if applicable. Not Applicable
   \item Information about consent from data providers/curators. Not Applicable
   \item Discussion of sensible content if applicable, e.g., personally identifiable information or offensive content. Not Applicable
 \end{enumerate}

 \item If you used crowdsourcing or conducted research with human subjects, check if you include:
 \begin{enumerate}
   \item The full text of instructions given to participants and screenshots. Not Applicable
   \item Descriptions of potential participant risks, with links to Institutional Review Board (IRB) approvals if applicable. Not Applicable
   \item The estimated hourly wage paid to participants and the total amount spent on participant compensation. Not Applicable
 \end{enumerate}

 \end{enumerate}

\end{document}